\documentclass[12pt]{article}

\def\bSig\mathbf{\Sigma}

\usepackage{amsmath}
\usepackage{amssymb}
\usepackage{natbib}
\usepackage[figuresright]{rotating}

\usepackage{multirow}

\begin{document}

\title{Bayesian Additive Adaptive Basis Tensor Product Models for Modeling
																	High Dimensional Surfaces: An application to high-throughput toxicity testing.}









\author
{Matthew W. Wheeler
Risk Analysis Branch\\ National Institute for Occupational Safety and Health,  Cincinnati, OH
}
\maketitle




\label{firstpage}


\begin{abstract}
Many modern data sets are sampled with error from complex high-dimensional surfaces. 
Methods such as tensor product splines or Gaussian processes are 
effective/well suited for characterizing a surface in two or three dimensions but may suffer from difficulties
when representing higher dimensional surfaces. Motivated by high throughput toxicity testing 
where observed dose-response curves are cross sections of a surface defined by a
chemical's structural properties, a model is developed to characterize this surface to predict
untested chemicals' dose-responses. This manuscript proposes a novel approach that 
models the multidimensional surface as a sum of learned basis functions
formed as the tensor product of lower dimensional functions, which are themselves representable
by a basis expansion learned from the data. The model is described, a Gibbs sampling algorithm proposed, and 
is investigated in a simulation study  as well as data taken from the US EPA's ToxCast high throughput toxicity 
testing platform.
\end{abstract}

%



\maketitle


%

\section{Introduction}
\label{s:intro}

Chemical toxicity testing is vital in determining the public
health hazards posed by chemical exposures. However, the number of chemicals
far outweighs the resources available to adequately test all chemicals,
which  leaves knowledge gaps when protecting public health.
For example, there are over $80,000$ chemicals in industrial use with
less than $600$ of these chemicals subject to long 
term \textit{in vivo} studies conducted by the 
National Toxicology Program, and most of these studies occur only
after significant public exposures. 

As an alternative to long term studies, which are expensive and take years
to complete, there has been an increased focus on the use of high throughput bioassays
to determine the toxicity of a given chemical. 
In an effort to understand the utility of these approaches for risk assessment,
many agencies have developed rich databases to study this problem. 
To such an end, the US EPA ToxCast chemical prioritization project 
\citep{judson2010} was created to collect dose-response information 
on thousands of chemicals for hundreds of \textit{in vitro} bioassays, and it
has been used to develop screening methods that prioritize chemicals for 
study based upon \textit{in vitro} toxicity.
Though these methods have shown utility in predicting toxicity for many chemicals
that pose a risk to the public health, there are many situations where 
the \textit{in vitro} bioassay information may not be available (e.g, 
a chemical may be so new that it has not been studied). In these cases,
it would be ideal if toxicity could be estimated \textit{in silico} by 
chemical structural activity relationship (SAR) information. 
Here the goal is to develop a model based on SAR information
that predicts the entire dose-response for a given assay. 
This manuscript is motivated by this problem.

 \subsection{Quantitative Structure Activity Relationships}

There is a large literature estimating chemical toxicity from 
SAR information. These approaches, termed Quantitative Structure Activity Relationships (QSAR) 
(for a recent review of the models and the statistical issues encountered see \citet{emmert2012}), estimate
a chemical's toxicity from the chemical's structural properties. 
Multiple linear regression has played a role in QSAR modeling since its inception\citep[p. 191]{roy2015}, but models where the predictor enters
into the relationship as a linear function often fail to describe the non-linear nature of the relationship. To address the non-linearity of
the predictor-response relationship, approaches such as neural networks 
\citep{devillers1996}, regression trees \citep{deconinck2005}, support vector 
machines \citep{czerminski2001,norinder2003}, and Gaussian processes \citep{burden2001} 
have been applied to the problem with varying levels of success. These approaches have been tailored to scalar responses, and, save one instance, have not been used to model the 
dose-response relationship, which may result in a loss of information. 

The only QSAR approach that has addressed the problem of estimating a dose-response curve is the work by \citet{low2015}. This approach defined a Bayesian regression tree over 
functions where the leaves of the tree represent a different dose-response surface. It was used to identify chemical properties related to the observed dose-response, and when this approach was applied to prediction the approach sometimes performed poorly in a
leave one out analysis. Further, it is computationally intensive 
and was estimated to take more than a week to analyze the data in motivating problem.

\subsection{Relevant Literature}
\begin{figure}
	\centering
		\includegraphics[width=0.85\textwidth]{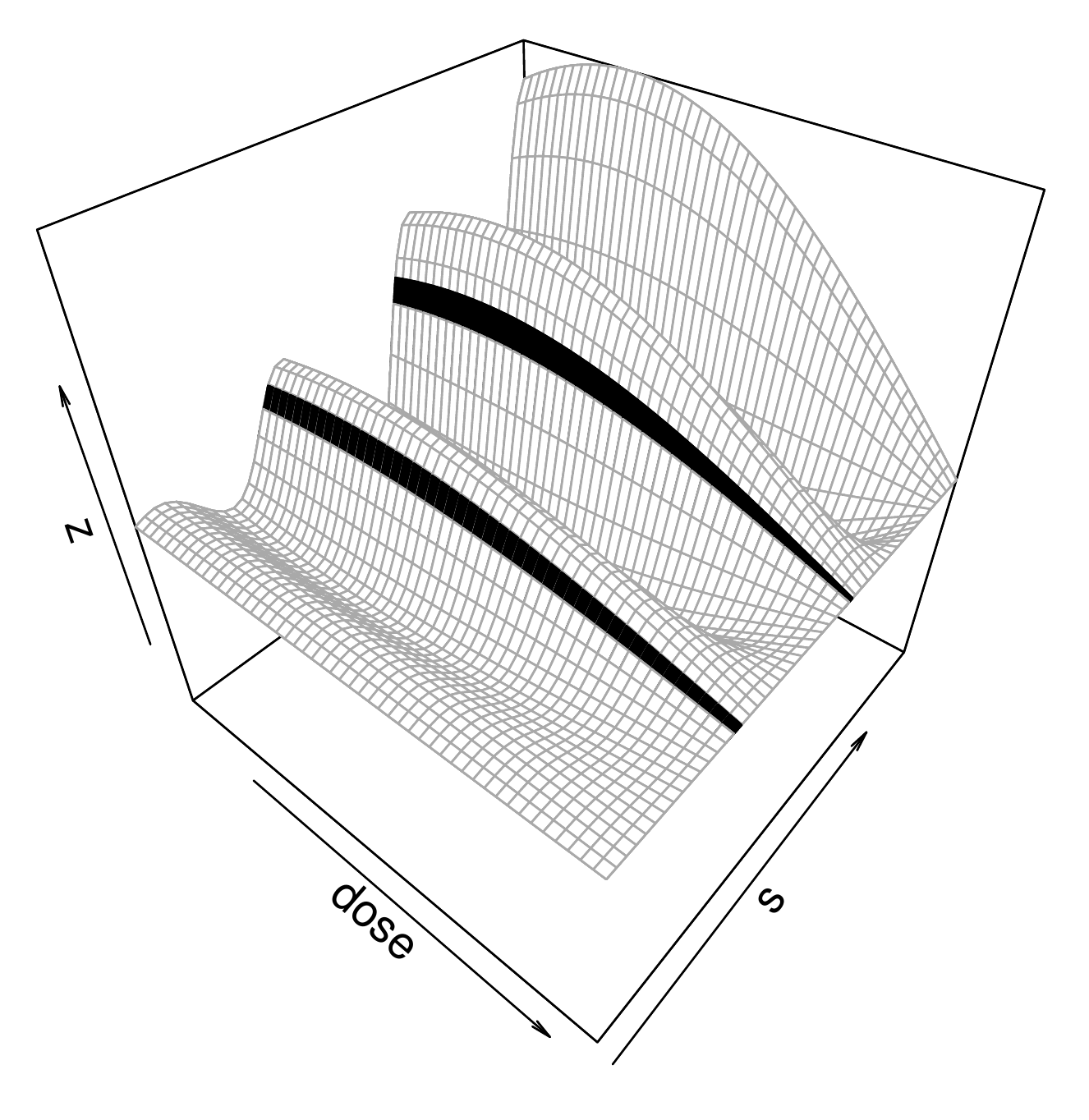}

	\caption{Example of the problem for a 2-dimensional surface.  Here two
	dose-response curves (black lines), which are cross sections of a larger surface, 
	are observed, and one is interested in this surface. }
		\label{fig:cross-sections}
\end{figure}

Assume that one obtains a $P$ dimensional vector $s \in \mathcal{S}$, and one wants to predict a $Q$ dimensional response over $d \in \mathcal{D}$  from
$s$.  Given  $s$ (e.g., SAR characteristics in the motivating
problem) and $d$ (e.g., doses in the motivating problem) one is interested in
 estimating an unknown $P+Q$ dimensional surface 
$h:(\mathcal{S} \times \mathcal{D}) \rightarrow \mathbb{R}$ where 
response curve $i$ is a cross section of $h(s,d)$ at $s_i.$ Figure (\ref{fig:cross-sections}) 
describes this in the case of the motivating example. 
Here two 1-dimensional cross sections (black lines) of a 2-dimensional surface are observed 
and one is interested the entire 2-dimensional surface. 

One may use a Gaussian process (GP, \citep{rasmussen2006} )
to characterize the entire $P+Q$ dimensional surface, but there are 
computational problems that make the use of a GP impractical. 
In the motivating example, there are over $4000$ unique $(s,d)$ pairs.   
GP regression requires 
inversion of the covariance matrix,  which is inherently an $\mathcal{O}(n^3)$ operation;
inverting a $4000 \times 4000$ matrix in each iteration of a Gibbs sampler is 
challenging, and, though the covariance matrix may be approximated 
leading to reduced computational burden (\citet{quinonero2005},\citet{banerjee2013}), it is the
author's experience that, in the higher dimensional case of the data example,
such approximations became accurate when the dimension of the approximation matrix approaches
that of the matrix it is approximating. This leads to minimal computational savings. 
Further, if the approximation approach can be used resulting in computational benefits, 
the the proposed method can be used in in conjunction with such approximations.

As an alternative to GPs, one can use tensor product 
splines (\citet{deBoor2000}, Chapter 17); 
however, even if one defines a functional basis having only two basis functions for each dimension,
the resulting tensor spline basis would have dimension $2^{P+Q},$ which  
is often computationally intractable. The proposed model sidesteps these issues by defining a
tensor product of learned multi-dimsensional surfaces defined on $\mathcal{S}$ and $\mathcal{D}.$

One could consider the problem from a functional data perspective \citep{ramsay2006,morris2015} 
by assuming a functional response over $\mathcal{S}\times\mathcal{D}.$ 
Approaches such as the functional linear array model
\citep{brockhaus2015} and  functional additive mixed models \citep{scheipl2015}
are closest to the proposed approach as they allow the surface to be
modeled as a basis defined on the entire space.  For high dimensional spaces where interactions
are appropriate, these approaches suffer the drawbacks because they do not allow for interactions
in all dimensions. Additionally, these approaches assume
that for any two observations $i,i'$ when  $s_i = s_{i'}$, loadings are drawn from a distribution 
independent of $s_{i}.$ This differs from the proposed approach assumes, which correlates loadings through $\mathcal{S}.$

Clustering the functional responses is also a possibility.
Here, one would model the surface using $\mathcal{D}$ and cluster using $\mathcal{S}.$
There are many functional clustering approaches
(see for example \citet{hall2001}, \citet{ferraty2006},\citet{zhang2010},\citet{sprechmann2010}
\citet{delaigle2012}, \citet{delaigle2013}, and references therein), but
these methods predict $s \in \mathcal{S}$ based upon observing the functional response $f(d)$, which 
is the opposite of the problem at hand. In this problem, one observes information on the group (i.e., $s$)
and one wishes to estimate the new response (i.e., $h(s,d)$).  However different, such approaches
can be seen as motivating the proposed method. 
Instead of clustering the loadings, the proposed approach assumes these loading for each
basis is a value on a continuous function over $\mathcal{S}.$ This is similar to clustering as similarity
is induced for any two responses having values in $\mathcal{S}$ that are sufficiently close. 

As the number of basis functions used in the model is choice that may impact the model's ability 
to represent an arbitrary surface, a sufficiently large number of basis functions are
included. Parsimony in this basis set is ensured by adapting to the number components in the model 
using the multiplicative gamma prior \citep{bhattacharya2011}.  This is a global shrinkage prior
is used to remove components from the sum by stochastically decreasing the prior variance 
of consecutive loadings in the sum to zero. The sum adapts to the actual 
number of basis functions needed to model the data and the choice in the 
number of basis functions is less important as long as the number is sufficiently large.

In what follows, section 2 defines the model. Section $3$
gives the data model for normal responses and outlines a sampling algorithm. 
Section 4 shows through a simulation study the method outperforms 
many traditional machine learning approaches, and section $5$ is
a data example applying the method to data from the US EPA's ToxCast
database. 

\section{Model}
\subsection{Basic Approach}
\label{s:model}
Consider modeling the surface $h: (\mathcal{S} \times \mathcal{D}) \rightarrow \mathbb{R}$
where $\mathcal{S} \subset \mathbb{R}^P,$ $ P \geq 1,$ and $\mathcal{D} \subset \mathbb{R}^Q,$ $Q \geq 1$. 
Given $s \in \mathcal{S}$ and $d \in \mathcal{D},$ tensor 
product spline approaches  \citep[chapter 17]{deBoor2000} approximate $h$ as a product of spline
functions defined over $\mathcal{S}$ and $\mathcal{D},$ i.e.,
\begin{align}
	g \otimes f &= g(s)f(d) \hspace{1cm} \forall s \in \mathcal{S}, d \in \mathcal{D}, \label{m:TP1}
\end{align}
for $g: \mathcal{S} \rightarrow \mathbb{R}^P$ and 
$f: \mathcal{D} \rightarrow \mathbb{R}^Q.$
The tensor product spline defines $g$ and $f$ to be in the span of a spline basis.
Assuming $g$ and $f$ are functions defined as
\begin{align*}
g(s) &= \sum_{j=1}^J \lambda_{j} \phi_{j}(s)
\end{align*}
and 
\begin{align*}
f(d) &= \sum_{l=1}^{L} \gamma_{l} \nu_{l}(d),
\end{align*}
where $\{\phi_{j}(s)\}_{j=1}^{J}$ and $\{\nu_j(d)\}_{l=1}^{L}$ are spline bases. The 
tensor product spline is 
\begin{align*}
g \otimes f = \sum_{j=1}^J \sum_{l=1}^{L} \rho_{jl}  \phi_{j}(s)\nu_{l}(d),
\end{align*}
where $\rho_{jl} = \lambda_{j} \gamma_{l}.$ As this approach typically uses 1- or 2-dimensional spline bases,
when the dimension of $(\mathcal{S} \times \mathcal{D})$ 
is large the tensor product becomes impractical as the number of functions in the 
tensor product increases exponentially.

Where tensor product spline models define the basis through a spline basis \textit{a priori},
many functional data approaches (e.g, \citet{montagna2012}) model functions from a basis learned from the data.   
Often the function space is not defined over $(\mathcal{S}\times \mathcal{D})$ directly, but is  
constructed on a smaller dimensional subspace, defined to be $\mathcal{D}$ 
in the present discussion. For cross section $i,$
this approach models $h(s_i,d),$  to be in the 
span of a finite basis $\{f_1(d),\ldots,f_K(d)\}.$  That is, 
\begin{align}
h(s_i,d) = \sum_{k=1}^{K} z_{ik} f_k(d), \label{m:eq1}
\end{align}
where $(z_{i1},\ldots,z_{iK})'$ is a vector of basis coefficients.
This effectively ignores $\mathcal{S}$ which may not be reasonable
in many applications.

To model $h(s,d)$ over $( \mathcal{S} \times \mathcal{D}),$ 
the functional data and tensor product approaches can be combined. 
The idea is to define a basis over $(\mathcal{S}\times \mathcal{D})$ where 
each basis function is the tensor product of two surfaces defined on $\mathcal{S}$
and $\mathcal{D}$ respectively.  Extending (\ref{m:eq1}), define $\{g_1(s),\ldots,g_K(s)\}$ to be surfaces on $\mathcal{S},$
and replace each $z_{ik}$ with $\zeta_k g_k(s_i).$
Now, loadings are continuous function indexed by $s$, and
a new basis $\{g_1\otimes f_1,\ldots, g_K \otimes f_K \}$ 
with 
\begin{align}
	h(s,d) &= \sum^{K}_{k=1} \zeta_k g_k(s) f_k(d) \label{S2:model}.
\end{align}

As pointed out by a referee, an alternative way to look at (\ref{S2:model}) is
that of an ensemble learner \citep{sollich1996},  which includes techniques such as
bagging \cite{breiman1996} and random forests \citep{breiman2001}. Ensemble learners describe 
the estimate as a weighted sum of learners (see \citet{murphy2012} and references therein). 
In this way, model (\ref{S2:model}) can be looked as a weighted sum  over tensor tensor product learners,
and related to an ensemble based approach.

To define a tensor product,  the functions 
 $\{g_k(s)\}_{k=1}^{K}$ and $\{f_k(d)\}_{k=1}^{K}$ 
must be in of a linear function space \citep[pg 293]{deboor2001}. 
For flexibility, let 
\begin{align*}
g_k &\sim \mathcal{GP}(0,\sigma^{g}_k(\cdot,\cdot)) \\
f_k &\sim \mathcal{GP}(0,\sigma^{f}_k(\cdot,\cdot))
\end{align*}
where $\sigma^{g}_k(\cdot,\cdot)$ and $\sigma^{f}_k(\cdot,\cdot)$ are positive
definite kernel  functions. This places  $g_k$ and $f_k,$ $k = 1,\ldots,  K,$
in a reproducing kernel Hilbert space defined by $\sigma^{g}_k(\cdot,\cdot)$ or 
$\sigma^{f}_k(\cdot,\cdot),$ and embeds
$h(s,d)$ in a linear space defined by the tensor product of these functions. 
By estimating the hyperparameters, this approach learns the basis of each 
function in (\ref{S2:model}), which may be preferable to a 
tensor spline approach that defines the basis \textit{a priori}. 

One may mistake this definition to be a GP defined by the tensor 
product of  covariance kernels (e.g., see \citet{bonilla2007}).
Such an approach forms a new covariance kernel over $(\mathcal{S} \times \mathcal{D})$ as a product 
of individual covariance kernels defined on $\mathcal{S}$ and $\mathcal{D}$.  For the proposed model,
the product is the modeled function and not the individual covariance kernels.

\subsection{Selection of K}

The number of elements in the basis is determined by $K.$
The larger $K$, the richer the class of functions the model can entertain. In 
many cases, one would not expect a large number of functions to 
 contribute to the surface and would prefer as few
components as possible. One could 
place a prior on $K,$ but it is difficult to find efficient sampling
algorithms in this case.  As an alternative, the multiplicative gamma process \citep{bhattacharya2011}
is used to define a prior over the $\zeta_1, \ldots \zeta_k$ that allows
the sum to adapt to the necessary number of components. 
Here, 
\begin{align*}
 \zeta_k \sim N\bigg(0, \bigg[\phi \prod_{j=1}^{k} \delta_j\bigg]^{-1} \bigg)
\end{align*}
with $\phi \sim \text{Ga}(1,1) $ and $\delta_j \sim \text{Ga}(a_1,1),$ $1 \leq j \leq K.$
This is an adaptive shrinkage prior over the functions. If 
$a_1 > 1,$ the variances are stochastically decreasing favoring
more shrinkage as $k$ increases.  The choice
of this prior implies the surface defined by $\zeta_k g_k(s)f_k(d)$ is
increasingly close to zero as $k$ increases. As many of the basis functions
contribute negligibly to modeling the surface, this 
induces effective basis selection.

\subsection{Relationships to other Models}
Though GPs are used in the specification of the model, one may use
alternative approaches such as polynomial spline models or process-convolution
approaches \citep{higdon2002}. Depending on the choice (\ref{S2:model}) can degenerate
into other methods.  For example, if  $\sigma_1^g,\ldots ,$ and $\sigma_K^g$ 
are defined as white noise processes and $f_1(s),\cdots,$ and $f_k(s)$ defined
to be in the linear span of the same finite basis,  the model is identical to the approach 
of \citet{montagna2012}, which connects the model to functional data approaches. 
In this way, the additive adaptive tensor product model can be looked
at as a functional model with loadings correlated 
by a continuous stochastic process over $\mathcal{S},$ instead of a white noise process.

If $P+Q=2$ and the functions $g_1$ and $f_1$ are defined using a spline basis,
this approach trivially degenerates to the tensor product spline model. Similarly, 
let each function in  $\{f_1(d),\ldots,f_k(d)\}$
be defined using a  common basis, with, 
\begin{align*}
	f_k(d) = \sum_{\ell=1}^{L} \beta_{\ell} \phi_\ell(d),
\end{align*}
where $\{\phi_\ell(d)\}_{\ell=1}^{L}$  is a  basis used for all $f_k(d)$. In this case, model  (\ref{S2:model})
can be re-written as 
\begin{align*}
	g(s,d) = \sum_{k=1}^{K} \sum_{\ell=1}^{L} \zeta_k \beta_{\ell} g_k(s) \phi_\ell(d).
\end{align*}
Letting $\beta^{\ast}_{lk} = \zeta_k \beta_{\ell}$, one arrives at a
tensor product model with learned basis $\{g_1(s),\ldots,g_{K}(s)\}$ and 
specified basis $\{\phi_\ell(d)\}_{\ell=1}^{L}.$

\subsection{Computational Benefits}
When $\mathcal{D}$ is  observed on a fixed number of points and 
$s$ is the same for each cross section, the proposed approach can deliver 
substantial reductions in the computational resources needed when compared to  
GP  regression.   
Let $r$ be the number of unique replicates on $\mathcal{D},$ and let
$n$ be the total number of observed cross sections. 
For a GP, the dimension of the corresponding covariance matrix is $rn.$ 
Inverting this matrix is an $\mathcal{O}( [rn]^3)$ operation.  For the proposed approach,
there are $K$ inversions of a matrix of dimension $r$ and $K$ inversions of a matrix of dimension
$n.$ This results in a computational complexity of $\mathcal{O}( K[r^3+n^3]),$ which can be
significantly less than a GP based method. In the data example this results in 
approximately $1/20th$ the resources needed as compared
to the GP approach, here $K=15$, $n=669$, and $r=7$. 
Savings increase as the experiment becomes more balanced. For
example, if $n=r$ and there are the same number of observations as in 
the data example, then a GP approach would require $10,000$ 
times more computing time than the proposed method. 

\section{Data Model and Sampling Algorithm}
\subsection{Data Model}
A Markov chain Monte Carlo (MCMC) sampling approach is outlined for normal errors.
Assume that for cross section  $i,$ $i = 1,\ldots, n,$ one observes $C_i$ measurements at
$\{(s_i,d_{ic})\}_{c=1}^{C_i}.$ For error prone observation
$y_{i}(s_i,d_{ic}),$ let 
\begin{align*}
	y_{i}(s_i,d_{ic}) = h(s_i,d_{ic}) + \epsilon_{ic},
\end{align*}
where $\epsilon_{ic} \sim N(0,\tau^{-1}).$  
Define  $N= \sum_{i=1}^{n} C_i.$
Model (\ref{S2:model}) assumes the surface is centered at zero;  here, let
the model be centered at $f_0(d).$

In defining $h_k(s)$ and $f_k(d)$, the covariance kernel, along with its
hyper-parameters, determines the smoothness of the function.
The squared exponential kernel is used to model smooth response surfaces.  Let 
\begin{align}
\sigma^{g}_k(s,s') = \varsigma_k \exp\bigg( -\theta_k \|s-s'\|^2\bigg) \label{S3:covmodel1}
\end{align}
and
\begin{align}
\sigma^{f}_k(d,d') =  \exp\bigg( -\omega_k \|d-d'\|^2 \bigg), \label{S3:covmodel2}
\end{align}
where $\| \cdot \|$ is the Euclidean norm, $\varsigma_k$ is
the prior variance, $\theta_k$ and $\omega_k$
are scale parameters, and $1 \leq k \leq K$. In this specification, the 
parameter $\zeta_k$ is not necessary; it is 
equivently defined through the variance of the GP 
where $\varsigma_k =  (\phi \prod_{j=1}^{k} \delta_j)^{-1}.$ To allow 
for a variance other than one for the process $f_0$, let 
$\sigma^{f}_0(d,d') = \nu \exp( -\omega_0 \|d-d'\|^2 ).$

Depending on the application any positive definite kernel
may be used.   For example,  
one may wish to use a Matern kernel, which 
is frequently used in spatial statistics. 
As $\mathcal{S}$  is described in the application as 
the distance between spatial locations in an abstract chemical 
space, such an approach may be appropriate.  Inpreliminary tests, 
a Matern kernel provides results that are qualitatively 
identical to the squared exponential kernel, and it is not considered
further.

Given these choices the data model is 
\begin{align}
	h(s,d) &= f_0(d) + \sum^{K}_{k=1} g_k(s) f_k(d). \label{S3:model}
\end{align}
Appropriate priors are placed 
over the hyperparameters of $\{\sigma_k^f\}_{k=0}^{K}$ and $\{\sigma_k^g\}_{k=1}^{K}.$ 
For the length parameter of the squared exponential kernels  in (\ref{S3:covmodel1}) and
(\ref{S3:covmodel2}), uniform distributions, i.e.,  $\text{Unif}(a,b),$ $0 < a < b,$ 
are placed over the scale parameters $\theta_k$ and 
$\omega_k.$ This places equal prior probability over a range of 
plausible values allowing the smoothness of the constituent basis to be learned.
The value $a$ is chosen so that correlations between any two points in the 
space of interest are close to $1,$ and $b$ is chosen so that correlation between any two points 
in the resultant correlation is approximately to $0.$

\subsection{Sampling Algorithm}

Define the vector of observations
$Y = \{y_1(s_1,d_{11}),\ldots,y_1(s_1,d_{1C_1}),$
$\ldots, y_1(s_n,d_{n1}),$$\ldots,y_1(s_1,d_{nC_n})\}'$ 
to be the $(N\times1)$ vector of measurements
across the $n$ observed curves. Likewise define
\begin{align*}
	\mathbf{G} &= 			\left[	 		\begin{array}{ccccc} 
																			1&g_1(s_1) & g_2(s_1) & \cdots &  g_K(s_1) \\
																			\vdots&\vdots   & \vdots   &        &  \vdots   \\
																			1&g_1(s_1) & g_2(s_1) & \cdots &  g_K(s_1) \\
																			\vdots&\vdots   & \vdots   &        &  \vdots   \\
																			1&g_1(s_n) & g_2(s_n) & \cdots &  g_K(s_n) \\
																			\vdots&\vdots   & \vdots   &        &  \vdots   \\
																			1&g_1(s_n) & g_2(s_n) & \cdots &  g_K(s_n) \\
																			\end{array} \right], 
\end{align*}
to be an $(N \times K+1)$ matrix, where each row corresponds to the 
loadings for observation $y_i(s_i,d_{ic}).$  
Let
\begin{align*}
	\mathbf{F} = \left[	 		\begin{array}{cccc} 
																			f_0(d_{11})& f_1(d_{11}) &  \cdots &  f_K(d_{11}) \\
																			\vdots     &\vdots   &         &  \vdots   \\
																			f_0(d_{1C_1})& f_1(d_{1C_1})  & \cdots &  f_K(d_{1C_1})\\
																			\vdots     &\vdots   &            &  \vdots   \\
																			f_0(d_{n1})& f_1(d_{n1}) & \cdots &  f_K(d_{n1})\\
																			\vdots     &\vdots     &        &  \vdots   \\
																			f_0(d_{nC_n})& f_1(d_{nC_n})  & \cdots &  f_K(d_{nC_n})\\
																			\end{array} \right] 
\end{align*}
be an $(N \times K+1)$ matrix, where each row represents the 
basis functions evaluated at $d_{ic}.$ 
Using these definitions, (\ref{S3:model}) is expressible  as 
\begin{align}
		Y = \bigg(\mathbf{G}\circ\mathbf{F}\bigg) J' + \mathbf{\epsilon}, \label{S3:sampf}
\end{align}
where $\circ$ is the Schur product, $J = (1,1,\ldots,1)$ is a $K+1$ row vector, and
$\mathbf{\epsilon} = (\epsilon_{11},\ldots,\epsilon_{iC_1},$
$\ldots,\epsilon_{n1},\ldots,\epsilon_{nC_n})'$ is the $(N \times 1)$ vector of 
error terms.  

Let $D = \{d^{\ast}_r\}_{r=1}^{R}$ be the set of $R$ uniquely observed inputs across all observations, and
define $\mathcal{I}^f$ to be an $(N \times R)$ matrix
where the rows corresponds to each element in $Y.$ For each row in $\mathcal{I}^f$,
all entries are set to zero except at column $r.$ This entry is
set to one, and it corresponds to the observation $y_i(s_i,d_{ic})$ such that $d^{\ast}_r  = d_{ic}.$
Likewise,  define the matrix $\mathcal{I}^{g}$
to be an $(N \times n)$ matrix.
For each row $r$, each entry is set to zero except at column $i,$
which is set to one.  This entry corresponds to 
$Y_r = y_i(s_i,d_{ic}).$ 

Sample from 
$\{g_k\}_{k=1}^K,$ $\{f_k\}_{k=0}^K,$ $\phi$, and $\{\delta_k\}_{k=1}^{K}$ in a series of 
Gibbs steps as follows: 

\begin{enumerate}
	\item For each $k$, $0 \leq k \leq K$ , 
		letting $Y^{\ast} = Y - (\mathbf{G}_{-k}\circ\mathbf{F}_{-k} ) J'_{-k},$
		where $G_{-k},$ $F_{-k},$ and $J_{-k}$ are $G,$ $F$ and $J$ 
		without column $k,$ sample $f_k \sim N(M,V)$ at
		$\{d^{\ast}_r\}_{r=1}^{R}.$  Here 
	\begin{align*}		
		V &= \Sigma_k(\tau\mathcal{G}'\mathcal{G}\Sigma_k + I)^{-1},\\
		M &= V (\tau \mathcal{G} Y^{\ast}),
	\end{align*}
		were $\Sigma_k$ is the $(R \times R)$ 
		covariance	function constructed from $\sigma^f_k(\cdot,\cdot),$ 
		$\mathcal{G}$ is an $(N \times R)$ matrix
		defined as $G_k \circ \mathcal{I}^f,$ and   $I$ is an
		$R \times R$ identity matrix. 

	\item For each $k$, $1 \leq k \leq K$ and letting $Y^{\ast} = Y - (\mathbf{G}_{-k}\circ\mathbf{F}_{-k} ) J'_{-k}$,
	 sample $g_k \sim N(M,V)$ at  $\{s_i\}_{i=1}^{n}.$ 
 	 Here 
	\begin{align*}
		V &= \Sigma_k(\tau\mathcal{F}'\mathcal{F}\Sigma_k + I)^{-1}, \\
		M &= V (\tau \mathcal{F}' Y^{\ast}),
	\end{align*} 
	where $\Sigma_k$ is the $(n \times n)$ covariance matrix formed
	from $\sigma^g_k(\cdot,\cdot)$,  $\mathcal{F}$ is
	an $(N \times n)$ matrix defined to be  $(FJ')\circ \mathcal{I}^g,$ and   $I$ is the
	$n \times n$ identity matrix.

	\item Sample $\phi \sim \mbox{Ga}(c,d)$, where
	\begin{align*}
			c &= K\frac{n}{2}+ a_1,\\
			d &= \bigg[\sum_{i=1}^K \bigg(\prod_{j=1}^i \delta_j\bigg) G_i^{'} \Sigma_i G_i  \bigg]+ 1,
	\end{align*}
	$G_i$ is a column vector from column $k$ of $G,$ and $\Sigma_i$ is the 
	matrix formed from $\exp(-\theta_i||s-s'||^2)$	
	\item For each $k$ sample $\delta_k \sim \mbox{Ga} (c,d),$ where
	\begin{align*}
			c &= (K-k+1)\frac{n}{2}+ a_1\\
			d &= \bigg[\sum_{i=k}^K  \phi \bigg(\prod_{j=1,j \neq k}^i  \delta_j\bigg) G_i^{'} \Sigma_i G_i\bigg] + 1
	\end{align*}
	$G_i$ is a column vector from column $i$ of $G,$ $\Sigma_i$ is the 
	matrix formed from $\exp(-\theta_i||s-s'||^2),$ and $\bigg(\prod_{j=1,j \neq 1}^1  \delta_j\bigg) = 1$. 
	
\end{enumerate}

The other parameters in the model are sampled using Gibbs or Metropolis
steps. The algorithm is written in the \texttt{R} programming language
\citep{r2015} using the \texttt{Rcpp}  \texttt{C++} 
extensions \citep{eddelbuettel2013} and the \texttt{RcppArmadillo} \citep{eddelbuettel2014}
linear algebra extensions and is available in the supplement.

When the number of unique observations is small, it is possible to 
develop a block Gibbs sampler for all of the $\{f_k\}_{k=0}^K$ and 
$\{g_k\}_{k=1}^{K}.$ 
In large problems, this requires the inversion of a large matrix
offsetting the benefits of the increased computational efficiency.

\subsection{Predictive Inference}

The posterior predictive distribution of $n^{\ast}$ unobserved cross sections 
of $h(s,d)$ at $\grave{S} = \{\grave{s}_1,\ldots,\grave{s}_{n^{\ast}}\}$ is
estimated through MCMC. Let the  
vector $g_k = (g_k(s_1),\ldots,g_k(s_n))'$ be observed at $S =(s_1,\ldots, s_n)'$ and one is interested
in  $\grave{g}_k = (g_k(\grave{s}_1),\ldots,g_k(\grave{s}_{n^{\ast}}))$
evaluated at $\grave{S}$; $g_k$ and $\grave{g}_k$
are jointly distributed as

\begin{align*}
	\left[ \begin{array}{c} g_k \\ \grave{g}_k \end{array} \right]
	\sim N\left(0,\left[ \begin{array}{cc} \Sigma^{g}_k(S,S) & \Sigma^{g}_k(S,\grave{S}) \\
																\Sigma^{g}_k(\grave{S},S) & \Sigma^{g}_k(\grave{S},\grave{S}) 
																\end{array} \right] \right).
\end{align*}
Here $\Sigma^{g}_k(S,S),$ $\Sigma^{g}_k(\grave{S},\grave{S}),$ $\Sigma^{g}_k(\grave{S},S)$
and $\Sigma^{g}_k(S,\grave{S})$ represent
covariance matrices given $S,$ $\grave{S}$ and $\sigma^{g}_k(\cdot,\cdot).$ Using properties
of the multivariate normal distribution, conditionally on $g_k$

\begin{align*}
\grave{g}_k \mid g_k \sim &N\bigg[\Sigma^{g}_k(\grave{S},S) \Sigma^{g}_k(S,S)^{-1} g_k,  \Sigma^{g}_k(\grave{S},\grave{S}) \\&- \Sigma^{g}_k(\grave{S},S) \Sigma^{g}_k(S,S)^{-1} \Sigma^{g}_k(S,\grave{S})\bigg]. 
\end{align*}

For each iteration, this expression is used to draw $\{\grave{g}_1,\ldots,\grave{g}_K\}.$
Given this draw, as well as $\{f_0,\ldots,f_K \}$ which represents $f_0(d),f_1(d)$ etc.,
evaluated at $D = \{d^{\ast}_r\}_{r=1}^R$, one can estimate the 
posterior predicted distribution of $h(\grave{S},D).$   If new 
$\grave{D}$ are of  interest, the same technique can be applied to estimating 
$\{\grave{f}_0,\ldots,\grave{f}_K.$ These values 
can be used with $\{\grave{g}_k\}_{k=1}^{K}$ to 
provide estimates for $h(\grave{S},\grave{D}).$

\section{Simulation}
This approach was tested on synthetic data. Here the dimension of $\mathcal{S}$ 
was chosen to be 2 or 3, and for a given dimension,  $50$ synthetic data sets were created, 
For each data set,  a total of $1000$ cross sections of $h(s,d)$ were observed at seven dose groups. 
Each data set contained a total of $7000$ observations. 

To create a data set,  the chemical information vector $s_i$, for chemical $i = 1, \ldots, 1000$, 
was sampled uniformly over the unit square/cube. 
At each $s_i,$ $h(s_i,d)$ was  sampled at $d= 0, 0.375, 0.75, 1.5,3.0,4.5,$ and $ 6,$ 
from 
\begin{align*}
	h(s_i,d) = \frac{\nu(s_i) d^m}{\kappa(s_i)^m+d^m},
\end{align*}
where $\nu(s_i)$ is the magnitude of the response and $\kappa(s_i)$ is
the dose $d$ where the response is at $50\%$ of the maximum. To vary 
the response over over  $\mathcal{S},$   a different zero centered  
Gaussian process, $z(s)$, was sampled 
at $\{s_i\}_{i=1}^{1000},$ and for each simulation;  
$\nu(s_i)$ and $\kappa(s_i)$ were functions of $z(s)$
with $\nu(s_i) = 11 \max( z(s_i),0)$ and $\kappa(s_i) = \max(4.5 - \nu(s_i),0).$
This resulted in the maximum response being between $0$ and approximately $50$ and 
the dose resulting in $50\%$ response being placed closer to zero for
steeper dose-responses. Sample data sets used in the simulation are available in 
the supplement. 

In specifying the model, priors were placed over parameters that reflect
assumptions based upon the smoothness of the curve. For $\{\sigma_k^g(\cdot,\cdot)\}_{k=1}^{K},$ 
each $\theta_k$ was drawn from a discrete uniform distribution
over the set $\{0.05,0.1,0.15,\ldots,4.05\}.$  Additionally,  
for $\{\sigma_k^f(\cdot,\cdot)\}_{k=0}^K,$ each $\omega_k \sim  \text{Unif}(0.1,1.5).$
For $\{\sigma_k^g(\cdot,\cdot)\}_{k=1}^K$,  $\delta_k \sim \text{Ga}(2,1),$ $1 \leq k \leq K$, which were the choices
used in \citep{bhattacharya2011}.  The choice of the parameters for the prior over the $\delta_k$ 
were examined, and the results were nearly identical with  $\delta_k \sim \text{Ga}(5,1).$ 
The prior specification for the model was completed by letting $\tau \sim \text{Ga}(1,1).$

A total of $12,000$ MCMC samples were taken with the first 
$2,000$ disregarded as burn in. Trace plots from multiple chains
were monitored for convergence. Convergence was fast usually occurring after 
$500$ iterations with approximately $4$ functions in the 
learned basis defining the surface (i.e. having posterior variance at or above
1).  

To analyze the choice of $K$, the performance of the model was 
evaluated for $K=1, 2, 3$ and $15.$  The estimates produced from these models 
were compared against bagged multivariate adaptive regression splines (MARS) \citep{friedman1991}
and bagged neural networks \citep{zhou2002} 
using the `caret' package in R \citep{caret2016}. Additionally, treed
Gaussian Processes \citep{gramacy2008} ,using the R `tgp' package \citep{gramacy2007},
were used in the comparison. All packages were run using their default settings;  $100$
bagged samples were used for both the MARS and neural network models. 
For the neural network model, $100$ hidden layers were used.  
Posterior predictions from the treed Gaussian process were obtained from the maximum
\textit{a posteriori} (MAP) estimate.  This was done as initial tests revealed estimates
sampled from the posterior distribution were no better than the MAP estimate, 
but sampling from the full posterior dramatically increased computation time making the full simulation
impossible.

For each data set, $N= 75,125,$ and $175$ curves were randomly sampled and used to 
train the respective model; the remaining curves were used as a hold out sample for posterior 
predictive inference with the true curve being compared against the predictions. 
All methods were compared using the mean squared predicted error (MSPE).

\begin{table}

		\caption{Mean squared prediction error in the simulation of the adaptive tensor product approach for four
		values of K as well as treed Gaussian processes, bagged neural networks, and bagged 
		multivariate regression splines (MARS).}\label{sim:tab1}
		\begin{tabular}{ccccccccc}
					&		& \multicolumn{4}{c}{Adaptive TP}&  &             &          \\
					&		&	K=1	  &	K=2	 & K=3 &	K=15	&	Neural Net	&	MARS  & Treed GP    \\
				\hline  \hline
\multirow{ 3}{*}{2-dimensions}	&     N=75		&	76.2	&	69.1 & 69.5	&	69.8	&	108.3	      &	$215.2^{1}$   &$839.7^{1}$  \\
				 			& N=125			&	56.9	&	48.5 & 48.8	&	48.7	&	92.4	      &	205.8         &$158.0^{1}$ \\
				 			& N=175			&	48.5	&	37.7 & 38.4	&	38.3	&	85.1	      &	198.8         & 61.1 \\
				\hline  \hline
\multirow{ 3}{*}{3-dimensions}	         & N=75		&	164.9	&	162.0       & 155.4 	&	155.4	&		185.2	      &$246.6^{1}$     &$1521.5^{1}$ \\
					 &N=125	&	128.6	&	125.0       & 	121.0	&	121.0	&		160.4       &	223.4              &$421.3^{1}$ \\
					 &N=175	&	106.3&  102.6       & 	 99.7	&	100.1	&		150.0	      &	217.7              &$163.5^{1}$ \\
				\multicolumn{7}{l}{\footnotesize$^{1}$ Trimmed mean used with $5\%$ of the upper and lower tails removed.} \\	
				\vspace{5mm}
		\end{tabular}

\end{table}

Table (\ref{sim:tab1}) describes the average mean squared predicted error across all simulations. For the
adaptive tensor product model there is an improvement in prediction when increasing $K.$ For the 2-dimensional case
the improvment occurs from $1$ to $2,$ but the results of $K=3$ and $15$ are almost identical to  $K=2.$ 
For the 3-dimensional case, improvements are seen up to $K=3$ with identical results for $K=15$. These results support
the assertion that one can make $K$ large and let the model adapt to the number of tensor products in the sum.

This table also gives the MSPE for the other methods. 
Compared to the other approaches, the adaptive tensor product approach is superior.
For $N=175$, the treed Gaussian process, which is often the closest competitor, produces mean square prediction errors that are
about $1.5$ times greater than the adaptive TP approach. For smaller values of $N,$ the treed GP as well 
as the bagged MARS approach failed to produce realistic predictions. In these cases, the 
trimmed mean with $5\%$ of the tails removed was used to provide a better estimate of center.

\begin{figure}
	\centerline{\includegraphics[width=0.85\textwidth]{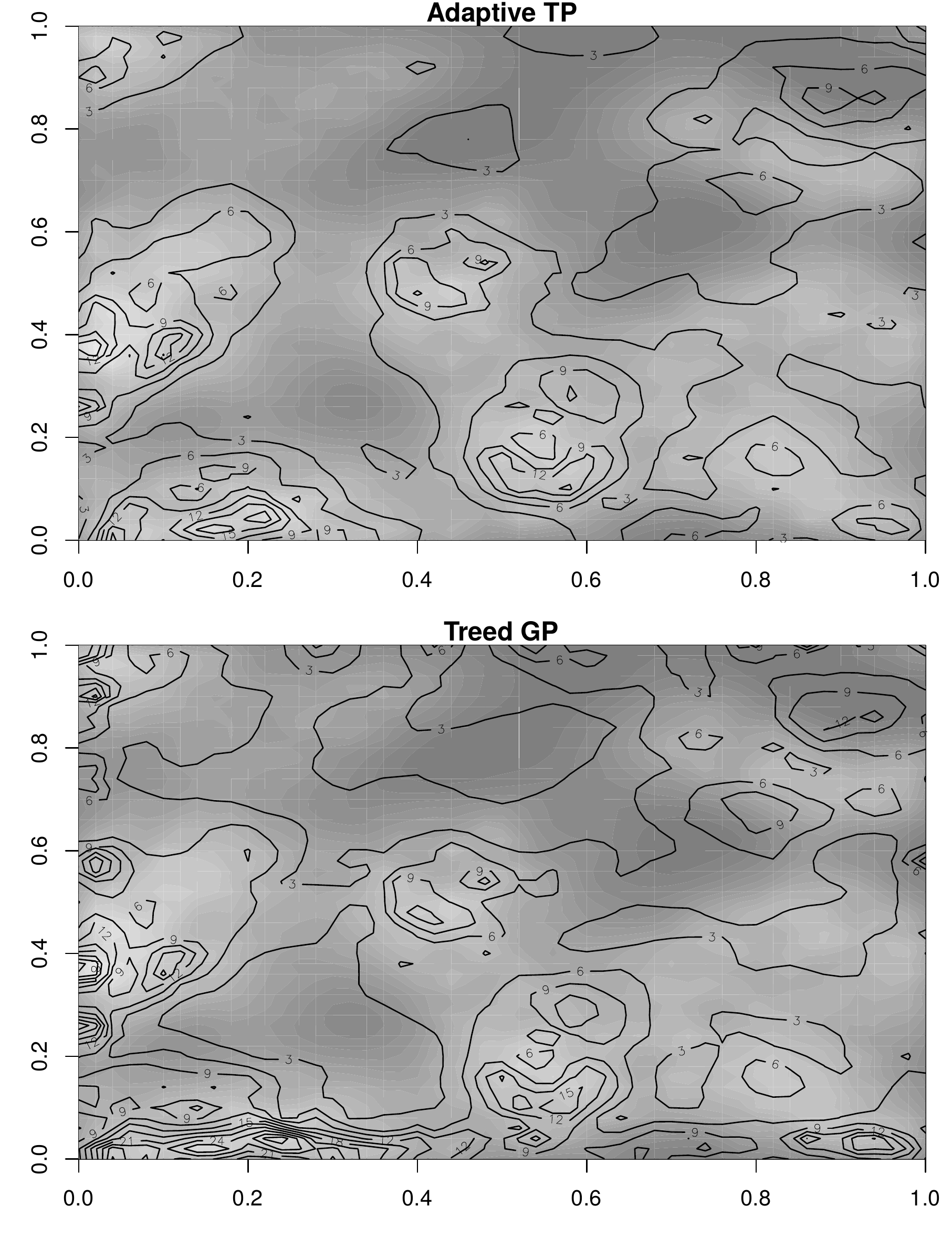}}
\caption{Comparison of the  predictive performance between the adaptive tensor product and the treed Gaussian process. 
In the figure, the corresponding model's  root mean squared predicted error is given as a contour plot. The  heat map represents the 
maximum dose response given the coordinate pair; lighter colors represent greater dose-response activity. }
\label{fig:sim}
\end{figure}

Figure (\ref{fig:sim}), shows where the gains in prediction take place. As a surrogate for the intensity of the true dose-response function, 
the intensity of the dose-response at $d=6$  across $\mathcal{S}$ is shown in the 2-dimensional case. 
Dark gray regions  are areas of little or no dose-response activity; lighter  
regions have the steeper dose-responses.  Contour
plots of the model's root MSPE are overlaid on the heat maps.  The top plot shows the performance 
of the adaptive TP and the bottom plot shows the performance of the treed GP, which was the closest competitor for this data set.
The figure shows  the adaptive TP is generally better at predicting the dose-response curve across $\mathcal{S}$
and that larger gains are made in regions of high dose response activity. This is most evident in the 
lower left corner of both plots, where regions of high dose response activity are located. 
At the peaks, the  root MSPE is almost double that for the treed Gaussian process.

\section{Data Example}
The approach is applied to data released from Phase II
of the ToxCast high throughput platform. The AttaGene PXR
assay was chosen as it has the highest number of active
dose-response curves across the chemicals tested. This assay targets 
the Preganene X receptor, which detects the presence of
foreign substances and up regulates proteins involved with detoxification in the body;
an increased response for this assay might relate to the relative toxicity of the given chemical.

Chemical descriptors were calculated using Mold$^2$ \citep{hong2008} where chemical 
structure was described from  simplified molecular-input line-entry system 
(SMILES) information \citep{weininger1988}. Mold$^2$ 
computes $777$ unique chemical descriptors.
For the set of  descriptors,  a principal component analysis 
was performed across all chemicals. This is a standard technique
in the QSAR liturature \citep[pg 44]{emmert2012}. Here, the first 
$38$ principal components, representing approximately $95\%$ of the 
descriptor variability,  were used as a surrogate for the chemical descriptor $s_i$.

The database was restricted to $969$ chemicals having SMILES information
available. In the assay, each chemical was tested across a range of doses between
$0\hspace{2pt} \mu M$ and $250\hspace{2pt} \mu M$ with no tests done at exactly a
zero dose. Eight doses were used per chemical, with each chemical being tested at
 different doses within the above range.  
Most chemicals had one observation per dose; however, some of the chemicals tested 
had  multiple observations per dose.  In total, the data set consisted of $9111$ 
data points from the $969$ distinct dose-response curves.

A random sample of $669$ chemicals was taken and the model wad trained to this data; in evaluating 
posterior predictive performance, the remaining $300$ chemicals were used as a hold out sample.
In this analysis, $d$ was the log dose, where
this value was rounded to two significant digits, and  the same prior specification was 
in the simulation was used to train the model. To compare the prediction
results, boosted MARS and neural networks were used; treed Gaussian processes 
were attempted, but the R package `tgp' crashed after 8 hours during burn-in.  The 
method of \citet{low2015} was also attempted,  but, the code was
designed such that each chemical is tested at the same doses with 
the same number of replications per dose point.  As the ToxCast data
are not in this format, the method could not be applied to the data. 

\begin{figure}
	
	\centerline{\includegraphics[width=0.85\textwidth]{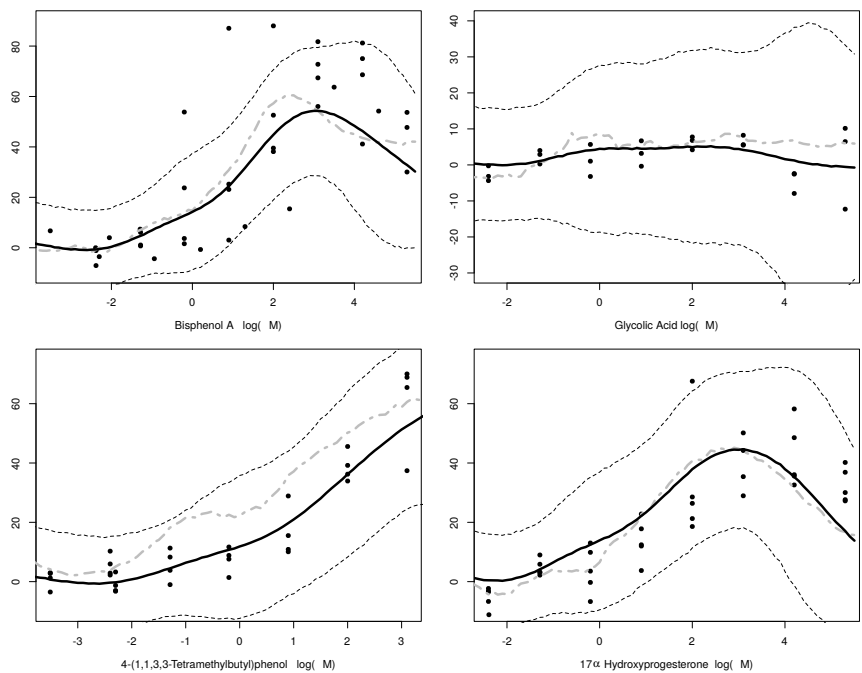} }
	\caption{Four posterior predicted dose-response curves (black line)
	with corresponding $90\%$ predicted credible intervals (dotted lines)
	for four chemicals in the hold out samples having repeated measurements per
	dose. Grey dash-dotted line represents the predicted response from the bagged 
	neural network.}
	\label{fig:4chem-pred-mult}	
\end{figure}

\begin{figure}
	\centerline{\includegraphics[width=0.85\textwidth]{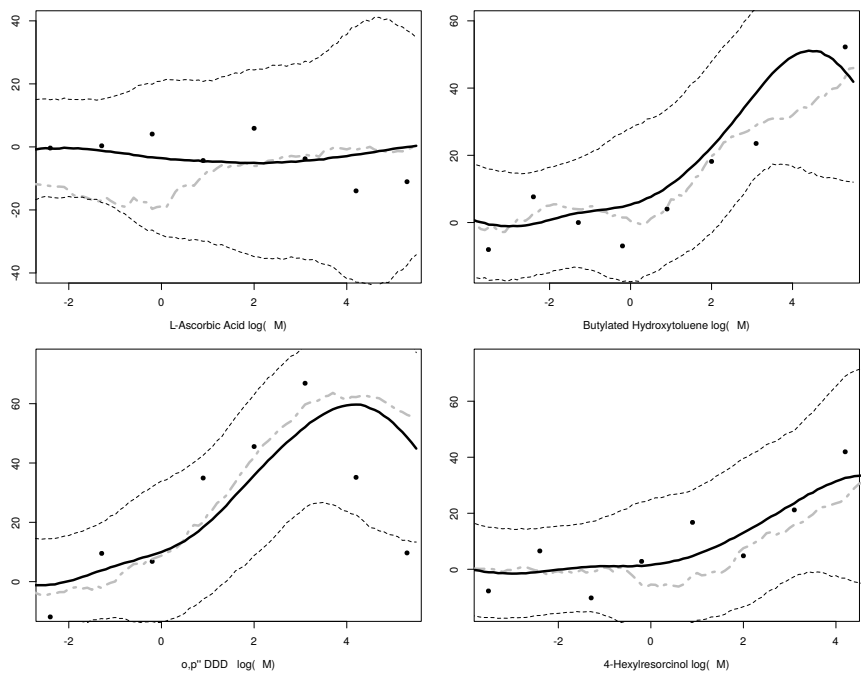} }
		\caption{Four posterior predicted dose-response curves (black line)
	with corresponding $90\%$ predicted credible intervals (dotted lines)
	for four chemicals in the hold out samples having repeated measurements per
	dose. Grey dash-dotted line represents the predicted response from the bagged 
	neural network.}
	\label{fig:4chem-pred-sing}
\end{figure}

Figures (\ref{fig:4chem-pred-mult}) and  (\ref{fig:4chem-pred-sing}) show
the posterior predicted curves (solid black line) with equal tail $90\%$ posterior
predicted quantiles (dashed line) for eight chemicals in the hold out
sample. Figure (\ref{fig:4chem-pred-mult}) describes the predictions
for chemicals having multiple measurements per dose group, and figure 
(\ref{fig:4chem-pred-sing}) gives predictions having only a single 
observation per dose group.  As compared to the observed data, 
these figures show the model provides accurate dose-response  
predictions across a variety of shapes and chemical profiles. Additionally
the grey dashed-dotted line represents the prediction using the bagged
neural network.  These estimates are frequently less smooth and further off
from the observed data than the adaptive tensor product splines.  
As additional confirmation the model is predicting dose-response curves
based upon the chemical information, one can look a the chemicals from 
a biological mode of action perspective.  For example in  (\ref{fig:4chem-pred-mult}), it is interesting
to note that the dose-response predictions for both Bisphenol A and 
$17\alpha$ Hydroxyprogesterone are similar,  because
both act similarly and are known to bind the estrogen receptor.  

In comparison to the other models, the adaptive tensor product approach also  had the lowest 
predicted mean squared  error and the predicted  mean absolute error for the data in the hold out sample. 
Here the model had a predicted mean squared error of $342.1$ and mean absolute error of $11.7$, as compared
to values of $354.7$ and $12.4$ for neural networks as well as $383.6$ and $13.4$ for MARS.  
These results are well in line with the simulation.  

One can also compare the ability of the posterior predictive distribution to predict the 
observations in the hold out sample. To do this,  
lower and upper tail cut-points defined by $p$ were estimated from the posterior 
predictive distribution.  The number of observations that were below the lower cut-point or above the
upper cut-point were counted. Under the assumption the posterior predictive distribution 
adequately describes the data, this count is $\mbox{Binomial}(2p,n)$ where $n$ equal to 
the number of observations for that chemical. To test this assumption, the $90\%$ critical value was computed
and compared to the count.  This was done for $p = 5, 10$ and $15\%;$ here, $88, 89,$ and 
$90\%$   of the posterior predictive distributions were at or below the $90\%$ critical 
value.  This supports the assertion that the model is accurately predicting the dose-response
relationship given the chemical descriptor information $\mathcal{S}$.

The method was also compared to standard QSAR approaches, which model a single data
point.  Similar to the standard QSAR methodology, a model was fit to each  data-set $i$
and the response associated with a given dose was computed to be observation $y_i$. 
This value was then used in the model 
\begin{align*}
y_i = x(s_i) + \epsilon_i, 
\end{align*}
where $x(s_i)$ is a Gaussian process with squared exponential covariance kernel.  
The model was fit on the same $669$ observations in the training set and predictions
were made of the response at that dose. Both this QSAR approach as well as the
adaptive tensor product approach were compared to the hold out samples
using the correlation coefficient, a standard practice in the QSAR literature. 
For the dose of $20 \mu M$ the standard QSAR approach had a correlation of $0.42$
as compared to the co-mixture approaches' correlation of $0.48.$ For the
dose of $100 \mu M$ a similar $0.06$ increase was seen, and, when observations
that have a chemical within the training set defined as `close' (i.e., a relative
distance between two chemicals less than 2.2), this improvement
in the correlation coefficient is almost $0.1$ (i.e, $0.5$ compared to $0.6$). 
This indicates that $10\%$ to $20\%$ improvements in the correlation coefficient
using the proposed approach.

\section{Conclusions}
The proposed approach allows one to model 
higher dimensional surfaces as a sum of a learned
basis, where the effective number of components in the basis
adapts to the complexity of the surface. 
In the simulation and the motivating problem, this method is 
shown to be superior to competing approaches, and, given the 
design of the experiment, it is shown to require significantly
less computational resources than GP approaches.  Though this approach is 
demonstrated for high throughput data, it is anticipated it can be used
for any multi-dimensional surface.

In terms of the  application, this model shows that 
dose-response curves can be estimated from chemical SAR information, which 
is a step forward in QSAR modeling. 
Though such an advance is useful investigating toxic effects, it can also be used
in therapeutic effects.  It is conceivable, that such an approach
can be used \textit{in silico} to find chemicals that have certain
therapeutic effects in certain pathways without eliciting toxic effects in other 
pathways. Such an approach may be of significant use
in drug development as well as chemical toxicity research.  

Future research may focus on extending this model to multi-output functional
responses. For example, multiple biossays dose-responses may be observed, and,
as they target similar pathways, are correlated. In such cases, it
may be reasonable to assume their responses are both correlated to each other
and related to the secondary input, which is the chemical used in the bioassay. 
Such an approach may allow for lower level  \textit{in vitro} bioassays, like the 
ToxCast endpoint studied here, to be used to model higher level \textit{in vivo} 
responses.
  




 \bibliographystyle{biom} 
 \bibliography{biblo}

\appendix


\label{lastpage}

\end{document}